\documentclass[twocolumn,10pt,letterpaper]{article}
\usepackage[top=2cm, bottom=2cm, left=1.8cm, right=1.8cm]{geometry}
\usepackage[utf8]{inputenc}

\usepackage{mathtools}
\usepackage{amsmath,amssymb,amsthm,amsfonts}
\usepackage{algorithm,algpseudocode}

\usepackage[usenames,dvipsnames]{xcolor}
\usepackage{array}
\usepackage{graphicx}
\usepackage{listings}
\usepackage{longtable}
\usepackage{float}
\usepackage{appendix}
\usepackage{textcomp}
\usepackage{caption}
\usepackage[retainorgcmds]{IEEEtrantools}
\usepackage{gensymb}
\usepackage{subcaption}
\usepackage{wrapfig}
\usepackage{placeins}
\usepackage{chngpage}
\usepackage{enumitem}
\usepackage[T1]{fontenc}
\usepackage{pgfplots}

\usepackage{multirow}

\usepackage{xcolor}

\definecolor{rouge_alerte}{RGB}{220,0,0} 
\definecolor{violet_doute}{RGB}{120,0,180} 

\def\targetInput{x_t}
\def\targetOutput{y_t}

\def\generator{G}
\def\discriminator{D}
\def\surrogate{S}
\def\targetModel{T}

\def\fidelity{F_S}
\def\accuracycomb{A_{S\circ G}}

\def\realdata{p_x}
\def\fakedata{p_z}

\def\cs{L_H}

\usepackage{tikz}
\tikzset{%
	>=latex,
            base/.style = {rectangle, rounded corners, draw=black,
                           minimum width=2.1cm, minimum height=0.6cm,
                           text centered, font=\sffamily},
  bluebox/.style = {base, fill=blue!30},
       redbox/.style = {base, fill=red!30},
    greenbox/.style = {base, fill=green!30},
    greybox/.style = {base, fill=black!10},
    blackbox/.style = {base, fill=black!30},
         orangebox/.style = {base, minimum width=2.5cm, fill=orange!15,
                           font=\ttfamily},
}
\usetikzlibrary{patterns}

\begin{document}

\title{\huge GAMIN: An Adversarial Approach to Black-Box Model Inversion} \author{Ulrich Aïvodji\thanks{UQAM, aivodji.ulrich@courrier.uqam.ca}, Sébastien Gambs\thanks{UQAM, gambs.sebastien@uqam.ca}, and Timon Ther\thanks{ISAE-SUPAERO, timon.ther@isae-alumni.net} \thanks{Corresponding Author}}

\date{}
\maketitle

\begin{abstract}
{Recent works have demonstrated that machine learning models are vulnerable to model inversion attacks, which lead to the exposure of sensitive information contained in their training dataset. 
While some model inversion attacks have been developed in the past in the black-box attack setting, in which the adversary does not have direct access to the structure of the model, few of these have been conducted so far against complex models such as deep neural networks.
In this paper, we introduce GAMIN (for \emph{Generative Adversarial Model INversion}), a new black-box model inversion attack framework achieving significant results even against deep models such as convolutional neural networks at a reasonable computing cost.
GAMIN is based on the continuous training of a surrogate model for the target model under attack and a generator whose objective is to generate inputs resembling those used to train the target model.
The attack was validated against various neural networks used as image classifiers. 
In particular, when attacking models trained on the MNIST dataset, GAMIN is able to extract recognizable digits for up to 60\% of labels produced by the target.
Attacks against skin classification models trained on the pilot parliament dataset also demonstrated the capacity to extract recognizable features from the targets.}
\end{abstract}

\section{Introduction}
\label{sec:introduction}

In recent years, the combination of increasing computing power, significant improvement in machine learning algorithms, as well as the augmentation of storage capacity have led to the development of prediction services. 
Those services rely on exploiting a wide variety of data, the collection of which has become an integral part of most digital and real-life markets. 
This push for personalized services has driven forward the use of sensitive data even in public-oriented applications such as evaluating the risk of recidivism of convicted inmates \cite{epic}. 
In addition, the gathered data has also become a vital asset for providers, which often seek to ensure the exclusivity over it.

Unfortunately, data-driven algorithms such as machine learning models are not immune to information leakage \cite{barreno06, kasiviswanathan12}.  
In particular, if those models have been trained over personal information, then an attack exploiting those leakages could result in a privacy breach (\emph{e.g.}, by inferring a sensitive feature for some of these individuals) for the individuals contained in the dataset used to train a learning algorithm. The potency of such an attack was first demonstrated in 2015 by Fredrikson, Jha and Ristenpart \cite{Fredrikson15}, extending as far as retrieving recognizable facial features or inferring with great accuracy a single sensitive feature. In subsequent works, efforts have been made to quantify the vulnerability of machine learning models against inference attacks related to privacy. For instance, new attacks, such as membership inference attack \cite{Shokri17}, as well as new metrics such as differential training privacy \cite{DTP}, have been developed.

Nonetheless some questions remain opened, especially with respect to the exploration of the possible attacks settings. 
Indeed, depending on the ability of the adversary to access the target model and his auxiliary knowledge of that model, the approach taken as well as the resulting efficiency can change significantly.
Considering \emph{model inversion attacks}, in which the attacker attempts to generate inputs resembling the original ones, Fredrikson and co-authors have left the design of effective black-box attacks as an open problem in their seminal paper \cite{Fredrikson15}. While a first significant step was proposed later \cite{Tramer16}, which consists in performing a black-box model extraction attack followed by the same white-box equation-solving attack used by Fredrikson and co-authors, this does not preclude the possibility that other approaches could be more efficient either in terms of the reconstruction accuracy or with respect to the knowledge required on the model attacked. In particular, one of the limits of the equation-solving attack is that it requires the knowledge of the target's architecture.

In this paper, we propose a novel approach that we coined as GAMIN (for \emph{Generative Adversarial Model INversion}) to specifically address the problem of black-box model inversion. Our attack is agnostic in the sense that we make no assumptions on the design of the target model or the original data distribution, which could arguably be the most difficult situation for the adversary \cite{sok-secu-privacy}. In a nutshell, our attack involves the simultaneous training of a \emph{surrogate model} approximating the decision boundaries of the target and a \emph{generator} that aims at producing the desired inputs, by mimicking the Generative Adversarial Networks (GANs) training process. To evaluate its cost and performance and how those are impacted by the target's architecture, our approach was tested against a collection of classifier models. To ensure the genericity of this evaluation, we chose to consider targets of various complexity, the design of which has been observed and tested in previous works in the literature. In particular, we have been able to demonstrate the potency of this attack against image classification models for digit recognition or skin color identification.

The outline of the paper is as follows. 
First in Section~\ref{sec:preliminaries}, we introduce the machine learning models necessary to the understanding of our work and their vulnerabilities with respect to various privacy attacks.
Afterwards in Section~\ref{sec:GAMIN}, we present the architecture of the GAMIN before describing in details how this attack performs against five models trained on the MNIST dataset in Section~\ref{sec:attacks}. Finally, we briefly review the related work in Section~\ref{sec:relatedWork} before concluding in Section~\ref{sec:conclusion} by discussing future work.

\section{Preliminaries}
\label{sec:preliminaries}

In this section, we review the relevant background on machine learning as well as model inversion attacks.

\subsection{Machine learning models}

\emph{Supervised learning} refers to the sub-domain of the machine learning field whose objective is to learn a function $F: X \rightarrow Y$ given a training dataset of input-output pairs $(x,y) \in X \times Y$.

The learning task is said to be a classification problem if $y$ is a label or a regression problem if $y$ is a continuous vector. 
The function $F$ is called a classifier when the possible values of $Y$ correspond to a finite number of classes and a regressor if the value of $Y$ is outputted from a continuous domain. 
A well-trained function $F$ should not only display a low error on the training set but also generalize to unknown $\tilde{x} \in X$.

To provide a concrete example, \emph{neural networks} have been recently re-popularized as a flexible solution for most supervised machine learning tasks. 
Neural networks rely on the composition of parametric functions to map an input $x$ to an output $y$ \cite{Lecun98}.
Those parametric functions correspond to layers of neurons, which are computing units parametrized by a weight vector $\theta$ and an activation function. 
Each unit, or neuron, outputs the result of its activation function to the weighted sum of its inputs. 
The weights themselves are updated by optimizers such as stochastic gradient descent with respect to a loss function.

\emph{Deep neural networks} are an extension of neural networks to more complex architectures.
Examples of such models include recurrent neural networks that allow for the treatment or production of sequences of inputs and outputs or convolutional layers that take into account the local context of each value in the vectors it processed \cite{AlexNet}.
By accounting for context and evolution, deep neural networks can extract and interpret patterns, which in turn allows for a substantial increase in performance \cite{vgg}. 
Such models are now widely used in fields such as computer vision or natural language processing.

\subsection{Generative adversarial networks}

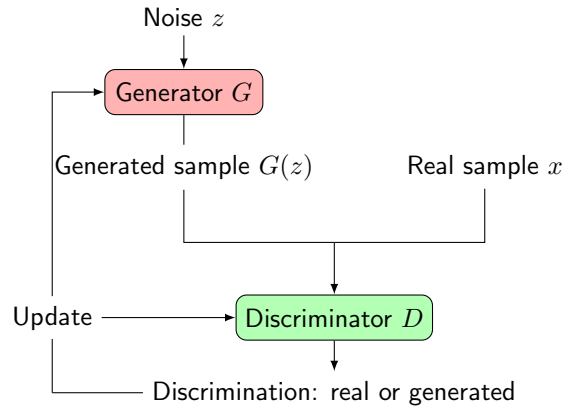
\begin{figure}[h!]
\centering
\begin{tikzpicture}[node distance=1.5cm, every node/.style={fill=white, font=\sffamily}, align=center]
\node (noiseBlock) [] {Noise $z$};
\node (generatorBlock) [redbox, below of=noiseBlock, yshift=0.5cm] {Generator $G$};
\node (genBlock) [below of=generatorBlock, yshift=0.5cm] {Generated sample $G(z)$};
\node (voidBlock) [right of=genBlock, xshift=0.5cm] {};
\node (actBlock) [right of=voidBlock, xshift=0.5cm] {Real sample $x$};
\node (voidBlock2) [below of=voidBlock,yshift=0.5cm] {};
\node (discBlock) [greenbox, below of=voidBlock2,yshift=0.5cm] {Discriminator $D$};
\node (resBlock) [below of=discBlock, yshift=0.5cm] {Discrimination: real or generated};
\node (voidBlock3) [left of=resBlock, xshift= -2.25cm] {};
\node (voidBlock4) [fill=white, above of=voidBlock3, yshift=-0.5cm] {Update};

\draw [->] (noiseBlock) -- (generatorBlock);
\draw (generatorBlock) -- (genBlock);
\draw (genBlock) |- (voidBlock2.center);
\draw (actBlock) |- (voidBlock2.center);
\draw [->] (voidBlock2.center) -- (discBlock);
\draw [->] (discBlock) -- (resBlock);
\draw (resBlock) -- (voidBlock3.center);
\draw (voidBlock3.center)  --  (voidBlock4);
\draw [->] (voidBlock4) -- (discBlock);
\draw [->] (voidBlock4) |- (generatorBlock);

\end{tikzpicture}
\centering
\caption{GAN architecture.} 
\label{fig:gan:schema}
\end{figure}

\emph{Generative Adversarial Networks} (GANs)~\cite{gan-Goodfellow2014} refers to a framework composed of two models competing against each other, hence the term ``adversarial''. 
In this setting, one of the models has the role of the generator $\generator{}$ while the other takes the role of the discriminator $\discriminator{}$. 
Figure~\ref{fig:gan:schema} depicts the associated learning process. 

In a nutshell, $\discriminator{}$ aims at distinguishing real data from fake one while $\generator{}$ tries to fools $\discriminator{}$ into taking a fake data it has generated for a real one. 
This process has been formally defined~\cite{gan-Goodfellow2014} as a minimax game between $\generator{}$ and $\discriminator{}$:
{\small
\begin{align*}
    \underset{\generator{}}{min} ~ \underset{\discriminator{}}{max} ~ \mathbb{E}_{x \sim \realdata{}(x)} [log(\discriminator{}(x))] + \mathbb{E}_{z \sim \fakedata{}(z)} [log(1 - \discriminator{}(\generator{}(z))],
\end{align*}
}
in which $\realdata{}(x)$ is the real data distribution and $\fakedata{}(z)$ is usually a simple distribution (\emph{e.g.}, $\mathcal{N}(0,1)$).

\subsection{Adversarial setting}

When attempting an attack on a machine learning model, an adversary can have different access and knowledge of the target model \cite{inversion-methodo} that we briefly introduced hereafter.

\emph{White-box setting.} In the white-box setting, the adversary knows the architecture of the target model as well as its parameters and internal states. 
Typically, this setting occurs when a malicious client obtains a complete trained model from a MLaaS (\emph{Machine Learning as a Service}) provider rather than a dataset and wishes to learn more about the original proprietary data this model has been trained on.

\emph{Black-box setting.} When only given black-box access, the adversary does not have a direct access to the model and can only query it with an input of his choice to receive the corresponding output (usually through an online API). 
For example, this scenario corresponds to the situation in which a company providing MLaaS is attacked by one of its users.
However, in some cases the adversary may have partial knowledge of the data used to train the target model (\emph{e.g.}, the characteristics of the distribution) or of the target model's architecture.

\emph{Black-box agnostic setting.}
The arguably hardest setting for the adversary is when it has absolutely no information on the training data (including no prior knowledge of the data distribution) as well as the model architecture or any of its parameters besides the dimensions of inputs and outputs. 
Basically, in this setting, which we refer to as the \emph{agnostic black-box setting}, the adversary can only query the target model to make inferences about it.

\subsection{Model inversion attacks against machine learning models}

\emph{Model inversion attacks}~\cite{fredrikson2014privacy,Fredrikson15} (also known as \emph{attribute inference attacks}~\cite{yeom2018privacy,jayaraman2019evaluating}) aim to infer hidden sensitive attribute of instances that belong to a particular output class of a target machine learning model. 
Pioneering works in this field~\cite{fredrikson2014privacy,Fredrikson15} have proposed a generic framework to conduct model inversion attacks that can be described as follows. 
Given a target model $T$, an instance $x$ --- whose non-sensitive attributes $x_1, x_2, \ldots x_{x_d-1}$ and attributes' prior probabilities are known to the adversary --- and the prediction $y=T(x)$ of the target model $T$, the attack identifies the value of the sensitive attribute $x_d$ maximizing the posterior probability $P(x_d | x_1, x_2, \ldots x_{x_d-1},y)$.

Model inversion attacks have been successfully conducted against a variety of models, including neural networks in the white-box setting with equation-solving methods~\cite{Fredrikson15}, through the reconstruction of recognizable portraits. 
The question of effective black-box attack with respect to this method was left as an open question. 
One of the proposed approach to realize this is to run a white-box inversion attack after a successful black-box model extraction attack~\cite{Tramer16}, thus achieving similar results at a fraction of the cost. 
However, none of these attacks were performed against a convolutional neural network or other types of deep neural networks. 
In addition, equations-solving attacks require prior knowledge of the target model's architecture, which severely undermines their feasibility in an agnostic black-box setting.

For image recognition tasks, the output of a model inversion attack is usually not a member of the training set, but rather an average of the features characterizing the class. 
Depending on the diversity of inputs yielding the targeted class, this average may lead to a more or less severe privacy breach. 
For instance, if a class correspond to a particular individual and the training instances corresponding to the latter have small variance (\emph{e.g.,} same location for the face or same shooting angle) then the result of the attack can help in re-identifying the individual. 
In contrast, if the output class characterizes highly diverse instances (\emph{e.g.,} car of very different types), then the output of the attack is more difficult to interpret. 
In this case, the output of the attack is very similar to that of a \emph{property inference attack}~\cite{ateniese2015hacking} (see Section~\ref{sec:inferences} for more details and examples about property inference attacks).

\section{Our approach: The GAMIN}
\label{sec:GAMIN}
In this section, we will first introduce the \emph{Generative Adversarial Model INversion} (GAMIN) framework before detailing the attack process.

\subsection{Overview}

Let us consider a target model $\targetModel{}$ and a label $\targetOutput{}$ corresponding to one of its labels. 
The objective of our attack is to characterize the inputs $\targetInput{}$ such that $\targetModel{}(\targetInput{}{})=\targetOutput{}$.

At a high level, GAMIN is composed of two deep neural networks, namely a \emph{generator} $\generator{}: Z \rightarrow X$ that maps noise $z \sim \mathcal{N}(0,1)$ to an input $x_G$ and a \emph{surrogate} model $\surrogate{}: X \rightarrow Y$ that outputs an estimation $\hat{y}$ of the target model's output. 
GAMIN allows to simultaneous train the surrogate model $\surrogate{}$ and the generator $\generator{}$ while performing a model inversion attack over $S$. 
Thus, the generator aims at learning the distribution of input $\targetInput{}$ associated to the label $\targetOutput{}$.

\begin{algorithm}[h!]
\caption{GAMIN training}\label{algo:gamin-train} 
\begin{algorithmic}
\Require{ ($G$, $\theta_G$) generator model, ($S$, $\theta_S$) shadow model, $\targetModel{}$ target model, $\targetOutput{}$ target label to invert, $k_0$ initial boundary-equilibrium factor, $\lambda_k, \gamma_k$ boundary-equilibrium update parameters }
\State {$k \leftarrow k_0$}
\For{$n$ epochs} 
\Statex \Comment {Generate artificial inputs from noise}
\State {$Z_G \sim \mathcal{N}(0,1)$}
\State {$X_G \leftarrow \generator{}(Z_G)$}
\Statex \Comment {Generate raw noise input}
\State {$X_S \sim \mathcal{N}(0,1)$}
\Statex \Comment {Query from the target model}
\State {$Y_G \leftarrow \targetModel{}(X_G)$}
\State {$Y_S \leftarrow \targetModel{}(X_S)$}
\Statex \Comment {Compute boundary-equilibrium loss and train surrogate}
\State{$L_S \leftarrow \cs(X_S,Y_S) - k * \cs(X_G,Y_G)$}
\State{$\theta_S \leftarrow train(X_S,Y_S,L_S)$}
\State{$k \leftarrow k + \lambda_k (\gamma_k \cs(X_S, Y_S) - \cs(X_G, Y_G) ) $}
\Statex \Comment {Train generator}
\State {$Z_G \sim \mathcal{N}(0,1)$}
\State {$\theta_{S \circ G} \leftarrow train(Z_G, y_t)$}
\EndFor
\end{algorithmic}
\end{algorithm}

Algorithm~\ref{algo:gamin-train} summarizes the training sequence of GAMIN.
For each step of the algorithm, a batch $Z_G$ is sampled from random noise $\mathcal{N}(0,1)$. 
Then, the generator uses the latter to produce a batch $X_G$. 
Afterwards, the target model $\targetModel{}$ is queried with (1) the output $X_G$ of the generator and (2) a batch $X_S$ sampled from random noise $\mathcal{N}(0,1)$. 
Subsequently, the predictions of $\targetModel{}$ on both $X_G$ and $X_S$ are used to compute the surrogate loss $L_\surrogate{}$ (see Section~\ref{began-loss}), which is then used to inform the training of the surrogate $\surrogate{}$. 
Finally, the generator's parameters are updated through the training of the combined model (see Section~\ref{combined:1}) $S \circ G$ on a batch $Z_G$ sampled from random noise $\mathcal{N}(0,1)$. 
After convergence, the surrogate learns the target model's decision boundaries and the generator learns to approximate $\targetInput{}$.  

Given the black-box agnostic setting of the adversary, the architecture of GAMIN has to be as generic as possible. More precisely, the only constraint put on the architecture is that the dimensions and ranges of the outputs of the generator match the inputs of the target model $\targetModel{}$ and GAMIN's surrogate $\surrogate{}$. 
We demonstrate this by using the same architecture --- for both GAMIN's surrogate and the generator --- for all the attacks conducted in this paper.
In addition, we have deliberately chosen for the surrogate model an architecture different from that of the target models to emphasize the absence of prior knowledge by the adversary. 
Technical details of GAMIN's surrogate and generator can be found respectively in Appendix~\ref{anx:archi:gamin:sur} and~\ref{anx:archi:gamin:gen}.

\subsection{Loss, metrics and convergence}
\label{metrics}

The adversarial nature of GAMIN requires the surrogate and the generator to be trained alternatively, each with its own losses and metrics. However, the training of one component impact the performances of the other. To overcome this challenge, we need to devise appropriate loss functions and metrics to have better control of the training process.

\subsubsection{Boundary-equilibrium loss for surrogate model}
\label{began-loss}
Initial experiments with this architecture proved its difficulty to assess convergence and to compare the performance performances of both the surrogate and the generator.  Since these issues are common in the GAN literature, we adapted the \emph{Boundary-Equilibrium Adaptive} loss from BEGAN~\cite{began} to GAMIN. The main idea is to have the loss function updating itself to reflect the trade-off between the surrogate loss on samples according to their nature.  More precisely, the distinction operated by the surrogate will be between noise inputs and generated samples. Further differences with the original BEGAN include adapting the setting from an auto-encoder perspective to our GAMIN setting. The loss for the surrogate model $L_S$ is defined as:
\begin{align*}
    L_S &= L_{H}(X_S, Y_S) - k_t * L_{H}(X_G, Y_G) \\
    k_{t+1} &= k_t + \lambda_k (\gamma_k L_{H}(X_S, Y_S) - L_{H}(X_G, Y_G) ),
\end{align*}
in which $L_{H}$ is the cross-entropy loss function, $X_S$ (resp. $X_G$) are noise inputs for the surrogate $\surrogate{}$ (resp. inputs crafted by the generator $\generator{}$), $Y_S$ (resp. $Y_G$) are predictions of the target model $\targetModel{}$ on $X_S$ (resp. $X_G$), $\lambda_k$ is a ``learning rate'' for the parameter $k_t$ and $\gamma_k$ is an equilibrium ratio objective. 

By allowing to balance the training of the surrogate and the generator, this loss helps in avoiding the issue of \emph{the unforgiving teacher}. This phenomenon is observed when a GAN's generator does not improve because its outputs are so far from expectations that it receives uniformly negative output~\cite{salimans}. In addition, the adaptive loss helps in determining and improving the component that is limiting the overall performance. Finally, it reduces the occurrences of \emph{catastrophic forgetting}, which may plague GAN-based architectures, and occurs when the learning of a new skill severely damages a previously learned one~\cite{french99}.

\subsubsection{Generator model training through combined networks}
\label{combined:1}

The update of the generator is achieved by merging both the generator and surrogate into a virtual ``combined'' neural network $S \circ G$. 
Then, the parameters of the combined network are updated through a standard training procedure. During this training, all the parameters related to the surrogate are kept frozen. 
Recall that the objective of the ``combined'' model is to map noise $z_G$ to the estimation of the target outputs upon receiving the generated inputs $S(G(z_G))$, to achieve $S(G(z_G)) \approx \targetOutput{} $ through training. 
Thus, the ``combined'' model is trained as performing a simple classification task with a cross-entropy loss function. 
Cross-entropy loss is defined as $L(y,\hat{y}) = - \sum_{i} y_i \textnormal{log}(\hat{y}_i)$, with $y$ being the distribution of probabilities across labels and $\hat{y}$ the estimation of these probabilities.

\subsubsection{Relevance metrics}

In this section, we discuss the metrics used to monitor and control the training and convergence of GAMIN.

\emph{Surrogate fidelity.}
\label{fidelity}
One of the fundamental measure for our approach is the \emph{surrogate fidelity}, which reflects the capacity of the surrogate to imitate the target model's behaviour. 
With the same input $x$ for the target model $\targetModel{}$ and the surrogate $\surrogate{}$, let $y = \targetModel{}(x)$ and $\hat{y} = \surrogate{}(x)$ be the outputs of these models. 
The fidelity $\fidelity{}$ of the surrogate can be expressed as $1 - \frac{||y - \hat{y}||}{||y||}$. 
In practice, we compute the mean of this fidelity, which is then $1-mae(y, \hat{y})$, $mae$ being the mean average error, and we test the fidelity over a batch of $64$ samples of uniform random noise. 
Note that fidelity differs from precision in that it needs not be evaluated over testing samples, which means that it can be measured during a black-box attack with no knowledge of the original data. A surrogate with high fidelity (up to $1$) estimates with good precision the decision boundaries of the target model.

\emph{Combined accuracy.} The generator's ability to craft inputs that are classified in the desired way by the surrogate is measured by the \emph{categorical accuracy} $\accuracycomb{}$ of the ``combined'' model (see Section \ref{combined:1}). 
When assessed in a batch, the categorical accuracy is the proportion of inputs categorized in the desired class.

\emph{BEGAN M-global score.}
\label{m-global}
The BEGAN-inspired adaptive loss presented in Section \ref{began-loss} comes with another useful metric from the same study: the \emph{global convergence score} $M_{global}$~\cite{began}.
This metric is defined as : $$ M_{global} = L_S + | \gamma_k L_S - L_{S\circ G} |,$$ in which $L_S$ is the loss of the surrogate on noise inputs and $L_{S\circ G}$ is the loss of the surrogate on inputs generated by the GAMIN generator.

The M-global convergence score provides an insight on the convergence of both networks, which is helpful to assess the overall convergence since the adversarial nature of the GAMIN architecture means an improvement of the surrogate may impact the generator (and vice versa). A strength of this score is that it can be computed in a black-box setting as it only relies on the discrimination between different types of inputs used to train the surrogate. 
This score and the adaptive loss are updated at the same time.

\subsubsection{Attack protocol and budget}
\label{protocol}

To perform an attack, GAMIN is given a query budget, which represents the number of queries that can be sent to the target model $\targetModel{}$.
GAMIN is then trained against the target model, saving both the surrogate and the generator whenever the M-global score improves (see Section \ref{m-global}) while tracking the remaining query budget. Upon exhausting the query budget, the generator with the lowest M-global score is used as the result of the attack to generate the targeted inputs.

This approach provides two main benefits. First, it helps to determine the actual number of queries needed to achieve the best result. Second, it preserves the best model observed during training, which means that catastrophic forgetting or a decrease in performance will have a lesser impact on the final performance of the attack.

\subsection{Post-processing}

Early experiments lead us to observe that when we attack datasets in which diverse inputs correspond to the same output (\emph{i.e}, the same class), the individual outputs of the generator do not converge to one of these specific inputs but rather that \emph{the mean of those generated inputs converged to the mean of the original inputs}.
Despite this limitation, we are nonetheless able to exploit the results of the GAMIN training by aggregating its results as described in the following procedure.

\subsubsection{Signal post-processing pipeline}\label{signal-pipeline}

Once the generator has been trained, the following process is used to retrieve the best version of the attacked class' mean. 
This process has to be applied separately for each channel when dealing with color images, each channel containing one color distribution.

\emph{Batch generation of outputs.} The first step consists in using our model to generate a large batch of outputs from noise. Our experiments have shown that $1000$ samples are sufficient to achieve a satisfactory sample of the variation between individual pictures.

\emph{Low-pass common frequency filter.} Then, each picture is transposed into the frequency domain via a Fast-Fourier Transform (FTT). The frequencies are then counted pixel-wise on all FTT images. 
Afterward, a filter is applied to remove frequencies that do not appear in at least $90\%$ of the FTT batch samples. 
Due to the frequency distribution on our experiment, this is equivalent to a low-pass filter.

\emph{Reconstruction.} The samples are mapped back to the spatial domain before applying a Gaussian blur to each of them, followed by an edge detection filter. Finally, the ``final'' output is obtained from the pixel-wise median of the transformed batch.

\subsubsection{Results}
\begin{figure}[h!]
\centering
\includegraphics[height=0.125\textwidth]{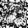} \hspace{0.025\textwidth} 
\includegraphics[height=0.125\textwidth]{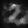} \hspace{0.025\textwidth} \includegraphics[height=0.125\textwidth]{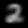}
\caption{Raw GAMIN output example (left), result of the post-processing on the same attack (center), and mean of the attacked class in original dataset (right).}
\label{fig:postprocess}
\end{figure}

The post-processing routine has considerably enhanced the visual quality of generated images by removing a good part of the noise cluttering the individual output (see Figure \ref{fig:postprocess} for an example).

Further post-processing such as thresholds or posterization could still be added to improve the quality of images, but those are not central to our study and are left to explore as future work.

\section{Black-box model inversion attacks}
\label{sec:attacks}

In this section, we present the results obtained against several target models. 
First, we present the datasets used to trained the target models as well as the models themselves before reporting on the results obtained. Afterwards, we discuss on how to evaluate the success of a GAMIN attack as well as the associated cost before finally reporting how models trained with differential privacy as a possible countermeasure are affected by the attack.
The source code of GAMIN is available on \emph{github}\footnote{https://github.com/definitively-not-a-lab-rat/gamin}.

\subsection{Targets and evaluation setting}

\subsubsection{MNIST dataset}

\begin{figure}[h!]
\centering
\includegraphics[width=0.4\textwidth]{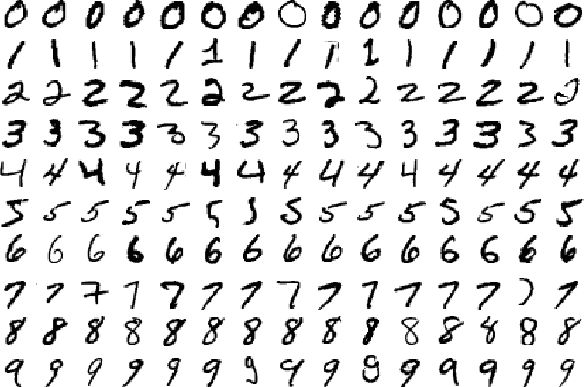}
\caption{Examples of MNIST digits.}\label{fig:mnistdigits}
\end{figure}

The MNIST Dataset of 70,000 handwritten digits (see examples in Figure \ref{fig:mnistdigits}) was used to train target models for a image recognition task. More precisely, in this case the recognition task consists in associating an image passed as input to a class corresponding to the digit the image represent. In this dataset, the images are of low-sizes (square of size 28 by 28 pixels) and black-and-white, which provides a relatively low-cost task.

The target models were trained on a defined proportion of the dataset accounting for 60,000 images, the rest being used for testing.

\subsubsection{Pilot parliament dataset}
\label{data:pilotparliament}

\begin{figure}[h!]
\centering
\includegraphics[width= 0.15\textwidth]{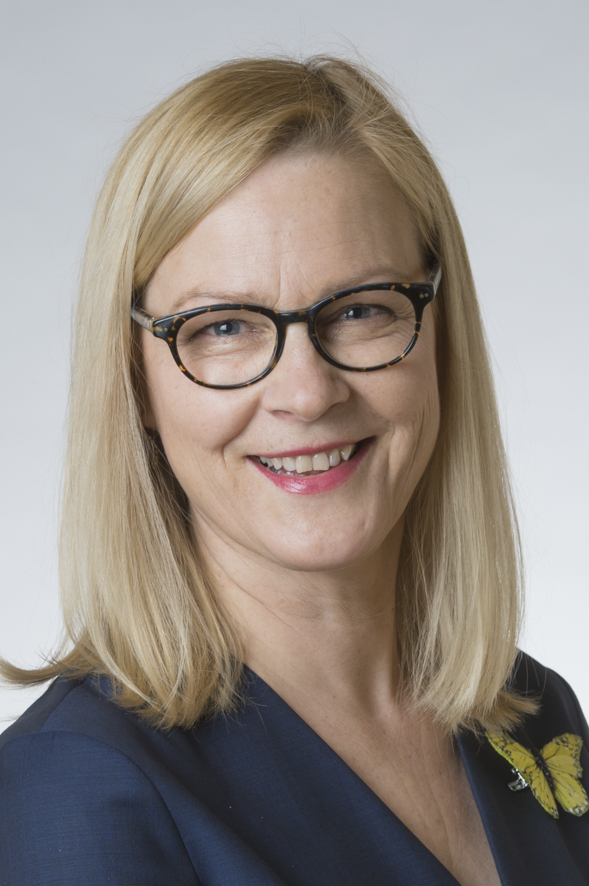} \hspace{0.05\textwidth}
\includegraphics[width= 0.15\textwidth]{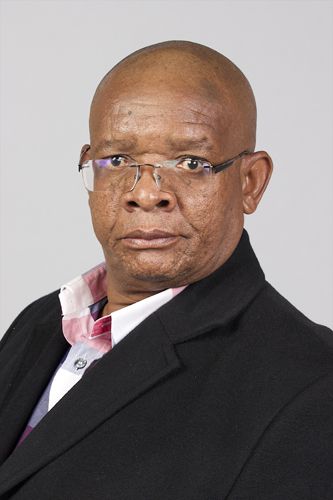}\hspace{0.05\textwidth} \\ \vspace{0.025\textwidth} 
\includegraphics[width= 0.15\textwidth]{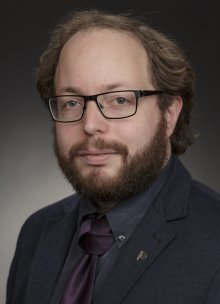}\hspace{0.05\textwidth}
\includegraphics[width= 0.15\textwidth]{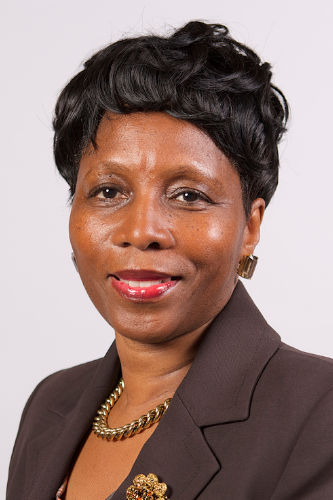}
\caption{Pilot parliament dataset samples.}
\end{figure}

Compiled by Joy Buolamwini \cite{cite-pilotparliament} from official portraits of parliament representatives across the globe, this dataset of 1270 color images of different sizes is classified according to sex and skin color (across the 6 Fitzpatrick skin categories), which are inherently sensitive attributes. 
In contrast to MNIST, this dataset is composed of colored pictures. 
Note that for our experiment we have reshape all images to the same size, which implies that some faces are distorted compared to the original versions. The target models were trained on 1079 portraits, the remainder being kept for the testing phase.

\subsubsection{Target models and training setup}

\emph{Types of models.}
Targets of various architectures were trained in order to see if a model's complexity impact the performance of the GAMIN attack. 
Overall, five models were considered as targets, which are named by order of increasing complexity: one \emph{softmax regression} (\emph{Soft}), two \emph{multilayer perceptron networks} (\emph{MLP1} and \emph{MLP2}), and two \emph{convolutional neural networks} (\emph{CNN1} and \emph{CNN2}). 
The architectures of these models are detailed in Appendix \ref{anx:targets}.
These architectures were chosen because of their prevalence in privacy and security-oriented literature. 
More precisely, Soft and MLP1 models were attacked in \cite{Fredrikson15} and \cite{Tramer16}, while MLP2, CNN1 and CNN2 were part of the targets in \cite{Papernot16}.

\emph{Performance with respect to overfitting.}
The objective of our study is to attack realistic models with imperfect performance and not ones that have overfitted.
Indeed, as overfitted models present very sharp decision boundaries, this usually renders the model inversion attack quite easy but also unrelated to real-world problems.
While training with the pilot parliament dataset, only \emph{CNN1} and \emph{CNN2} models delivered good results.
However, the \emph{Soft} model was also included in the study for comparison despite its lesser performance.
The full details of each target model performance are summarized in Appendix \ref{anx:target-performance}.

\emph{Training setup.}
The surrogate has been trained with an Adam optimizer \cite{adam} regarding to this adaptive loss with the following parameters for the optimizer: learning rate of $10^{-5}$, $\beta_1 = 0.99$, $\beta_2 = 0.99$, $\epsilon=10^{-8}$ and for the adaptive loss: $\lambda_k = 0.01$, $\gamma_k$ = $0.5$ and $k_0 = 0.001$. 
The generator has also been trained with an Adam optimizer configured similarly to the surrogate's with the following parameters for the optimizer: learning rate of $10^{-5}$, $\beta_1 = 0.99$, $\beta_2 = 0.99$ and $\epsilon =10^{-8}$.
These parameters are fairly standard and yield good results for the problems the GAMIN was confronted to.
Remark that an extensive hyperparameter search is a non-trivial process in the black-box setting.

\subsection{Attack against MNIST-trained targets} \label{attack:results}

\subsubsection{Evaluation survey}
To measure how well GAMIN solves the model inversion problem from a privacy point of view, we cannot rely on pixel-wise errors or distances between distributions. 
Indeed, since the objective is to extract sensitive information from a model, we have to rely on human interpretation of the results to evaluate whether they successfully disclose information about the training data or not. 
In order to measure this information leakage on MNIST, we set up a survey in which we showed to $13$ participants the results of the model inversion of the GAMIN and asked them to guess which specific digit they thought was represented. 

\subsubsection{Results} 
\label{attack:results:results}
\begin{figure}[h!]
\centering
\includegraphics[width=0.5\textwidth]{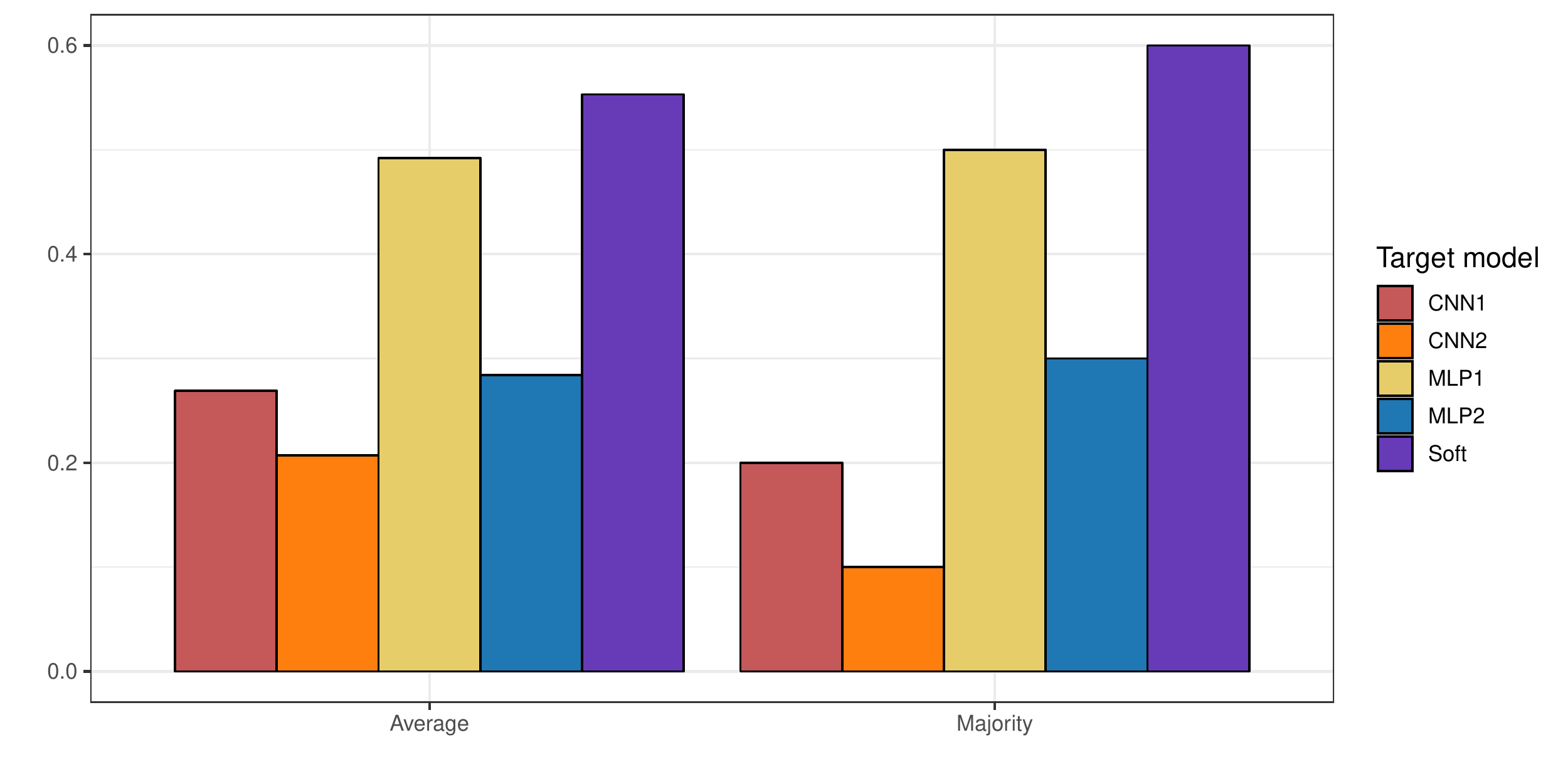}
\caption{Results of the survey for MNIST-based models attack. For each target model, the average (over all the target digits) of the proportion of correct identification is given as well as the proportion of target digits with at least $50\%$ of correct identification.}
\label{fig:result:mnist:survey1}
\end{figure}

The results of the survey are summarized in Figure \ref{fig:result:mnist:survey1}. 
In this figure, \emph{average} displays the percentage of correct inference of digits averaged across all $10$ digits while \emph{majority} depicts the percentage of classes (\emph{i.e.}, digits) for which more than half the volunteers guessed the correct digit from the pictures they were shown. 
In our experiments, a \emph{correct} guess is awarded only when the participant picked the right digit. 
In opposition, giving a wrong answer or selecting the option ``unable to decide'' answers were counted as negative results.

In the best case, when attacking the soft model, volunteers were able to pick the attacked digit in $55\%$ of cases, assessing the reconstruction of $6$ out of $10$ digits. 
However, the ability of the GAMIN to invert models seems to be dramatically affected by the complexity of the target model, with both average number of correct answers and number of classes successfully inverted dropping to respectively $20\%$ and $1$ out of $10$ for the CNN2 architecture.

Convolutional models, which were not targeted by model inversion attacks in previous works, proved to be more resistant to these attacks. 
In particular, the drop in reconstruction accuracy that occurs when attacking deeper models could be explained directly by their complexity, as the link between decision boundaries and classification becomes less and less direct.

\begin{figure}[h!]
\centering
\hspace{0.02\textwidth} Soft \hfill
MLP1 \hfill
MLP2 \hfill
CNN1 \hfill
CNN2 \hspace{0.01\textwidth}

\includegraphics[width=0.08\textwidth]{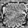} \hspace{0.01\textwidth}
\includegraphics[width=0.08\textwidth]{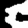} \hspace{0.01\textwidth}
\includegraphics[width=0.08\textwidth]{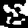} \hspace{0.01\textwidth}
\includegraphics[width=0.08\textwidth]{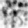} \hspace{0.01\textwidth}
\includegraphics[width=0.08\textwidth]{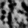}

\vspace{0.01\textwidth}

\includegraphics[width=0.08\textwidth]{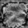} \hspace{0.01\textwidth}
\includegraphics[width=0.08\textwidth]{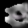} \hspace{0.01\textwidth}
\includegraphics[width=0.08\textwidth]{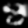} \hspace{0.01\textwidth}
\includegraphics[width=0.08\textwidth]{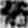} \hspace{0.01\textwidth}
\includegraphics[width=0.08\textwidth]{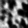}

\caption{Reconstruction of the same digit (\emph{i.e.}, eight) issued from attacking different models. Raw outputs samples are located on the top while postprocessed outputs can be found on the bottom).} \label{fig:result:eights}
\end{figure}

By taking a look at the GAMIN output for the same digit, we can observe the captured decision boundaries (see Figure~\ref{fig:result:eights}), which confirm that the reconstruction given when attacking the soft architecture has usually sharp and identifiable features that are directly linked to the value of a region, whereas models using convolutions (\emph{i.e.}, CNN1 and CNN2) use patterns and filters before selecting the class resulting in blurred and more abstract images.

\begin{figure}[h!]
\centering
\includegraphics[width=0.5\textwidth]{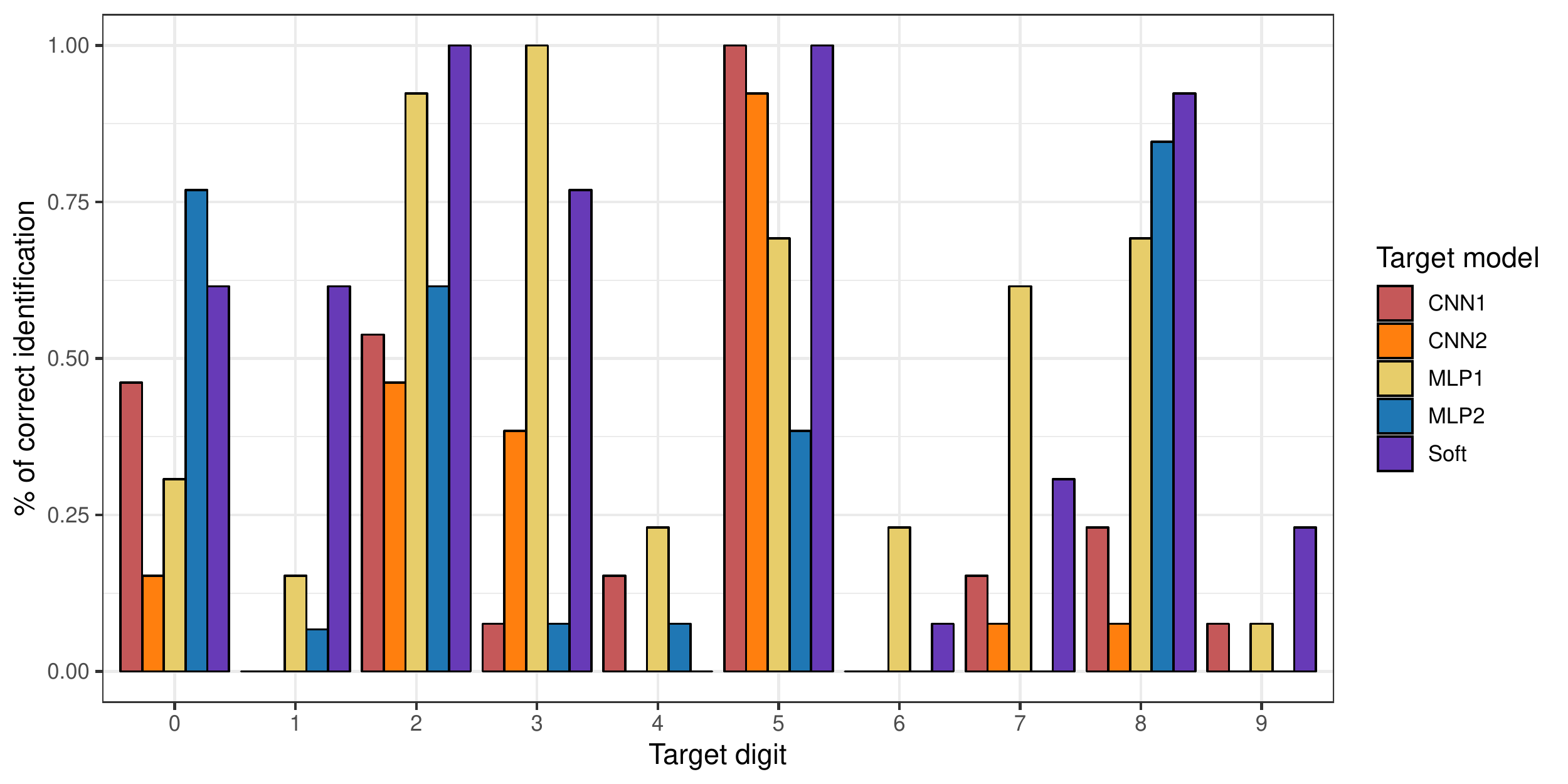}
\caption{Results of the survey for MNIST-based models attack. 
For each target model and for each target digit, the proportion of correct identification is given.}
\label{fig:result:mnist:survey2}
\end{figure}

A closer look at the results breakdown between digits (displayed on Figure \ref{fig:result:mnist:survey2}) shows a strong disparity among digits. 
Interestingly, some digits ($2$ and $3$) are consistently inverted with success, whereas others such as for instance $1$, display the opposite behaviour as the GAMIN attack fails to reconstruct these most of the time. 
A plausible explanation of the difference in accuracy when attacking those different classes could be the amount of recognizable features each of these class have. 
While this might considered a subjective notion, recall that the mean of the GAMIN outputs tends to converge to the mean of the targeted class.

\subsection{Metrics and relation to success of the attack}

The average metrics defined previously in Section \ref{metrics} for each target model trained on MNIST are displayed in Table \ref{tab:results:metrics}. 
For each GAMIN attack, the average over all $10$ digits is denoted as $\fidelity{}$ for the fidelity of the surrogate model, $\accuracycomb{}$ for the categorical accuracy of the combined model and $M_{global}$ for the best global score of convergence. 
The ratio of correct re-identification by the respondants of the survey is also provided for comparison.

\begin{figure}[h!]
\centering
\includegraphics[width = 0.08\textwidth]{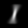} \hspace{0.04\textwidth}
\includegraphics[width = 0.08\textwidth]{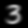}
\caption{Averages for digits 1 and 3 on the MNIST dataset.} \label{fig:means13}
\end{figure}

First, we can observe that there is an absence of correlations between the accuracy $\accuracycomb{}$ of the combined model and the success rate of the re-identification by human volunteers. 
For instance, by comparing the mean of a digit with frequent success (the 3) to a digit with rare success (such as the 1) displayed on Figure \ref{fig:means13}, we observe that the classification task of a representation of an image's mean might not be sufficient to evaluate completely the success of the model inversion attack.
Nonetheless, the relatively high fidelity scores yielded by most attacks demonstrate that the surrogate is able to capture, at least partially, the decision boundaries of the target model. 

\begin{table}[h!]
\centering
$\begin{array}{ |*{4}{c|}|c| }
\hline 
\textnormal{Target model}  & \fidelity{}  & \accuracycomb{}  & M_{global} & \%\text{correct}    \\ \hline \hline
\textnormal{Soft}          & 0.91 & 0.98 & 0.06       & 0.553  \\ \hline
\textnormal{MLP1}          & 0.73 & 1.0  & 0.08       & 0.492  \\ \hline
\textnormal{MLP2}          & 0.71 & 1.0  & 0.18       & 0.284  \\ \hline
\textnormal{CNN1}          & 0.83 & 0.74 & 0.19       & 0.269  \\ \hline
\textnormal{CNN2}          & 0.56 & 1.0  & 0.16       & 0.207  \\ \hline
\end{array}$
\centering
\caption{Average performances across all digits of inversion attacks against the target models. $\fidelity{}$, $\accuracycomb{}$, $M_{global}$  and $\%$\text{correct} refer respectively to the fidelity of the surrogate, the categorical accuracy of the combined model, the global convergence score and the accuracy as evaluated through the survey.}\label{tab:results:metrics}
\end{table}

To further investigate the link between $\fidelity{}$ and $M_{global}$ as well as the attack success, we provide an extensive review of the final score when attacking each digit against the MLP1 architecture in Table \ref{tab:results:mlp1metrics}.
These results show that a surrogate model achieving a very high fidelity or very low global convergence score might not be sufficient to achieve a good model inversion. 
In particular, this table provides examples with digits $1$ and $9$ of attacks that failed to reconstruct recognizable digits, yet achieve $M_{global}$ scores among the lowest. 
Furthermore, attacks against digit $1$ also achieved fidelity scores comparable to that of any successful attack.

\begin{table}[h!]
\centering
$\begin{array}{ |*{3}{c|}|c| }
\hline 
\textnormal{Target digit}  & \fidelity{}   & M_{global} & \%\text{correct}   \\ \hline \hline
\textnormal{0}             & 0.83  & 0.06       & 0.31  \\ \hline
\textnormal{1}             & 0.75  & 0.07       & 0.15   \\ \hline
\textnormal{2}             & 0.78  & 0.07       & 0.92  \\ \hline
\textnormal{3}             & 0.74  & 0.09       & 1.00   \\ \hline
\textnormal{4}             & 0.73  & 0.10       & 0.23  \\ \hline
\textnormal{5}             & 0.75  & 0.11       & 0.69  \\ \hline
\textnormal{6}             & 0.77  & 0.07       & 0.23  \\ \hline
\textnormal{7}             & 0.86  & 0.09       & 0.62  \\ \hline
\textnormal{8}             & 0.77  & 0.08       & 0.69  \\ \hline
\textnormal{9}             & 0.36  & 0.08       & 0.08  \\ \hline
\end{array}$
\centering
\caption{Per-digit performances of inversion attacks against the MLP1 model. $\fidelity{}$, $M_{global}$,  and $\%$\text{correct} are respectively the fidelity of the surrogate, the global convergence score, and the accuracy as evaluated with the survey}\label{tab:results:mlp1metrics}
\end{table}

The results demonstrate that the GAMIN surrogate needs to achieve a high fidelity $\fidelity{}$ to succeed, which is consistent with using fidelity as a measure of how well the decision boundaries are captured. 
However, this is a necessary but not a sufficient condition to ensure the success of the attack.

\subsection{Attacks against skin color classifiers}

\begin{figure}[h!]
\centering
\includegraphics[width= 0.15\textwidth]{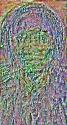} \hspace{0.05\textwidth}
\includegraphics[width= 0.15\textwidth]{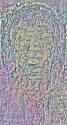} \\ 
\caption{Outputs of attack against a CNN1 skin color classifier.} \label{fig:gamin-pp-skin-cnn1}
\end{figure}

\begin{figure}[h!]
\centering
\includegraphics[width= 0.15\textwidth]{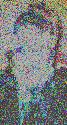} \hspace{0.05\textwidth}
\includegraphics[width= 0.15\textwidth]{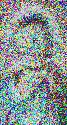} \\ \caption{Outputs of attack against a soft skin color classifier.} \label{fig:gamin-pp-skin-soft}
\end{figure}

Subtle, yet recognizable facial shapes can be observed on the images generated after attacking a skin color classifier trained on the pilot parliament dataset (see Figures \ref{fig:gamin-pp-skin-cnn1} and \ref{fig:gamin-pp-skin-soft}).
Interestingly enough, even the central area of the image does not give a hint of the skin color of individuals. 
Instead, the decision boundaries captured by the surrogate model are based on contours.

\begin{figure}[h!]
\centering
\includegraphics[width= 0.15\textwidth]{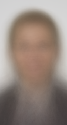} \hspace{0.05\textwidth}
\includegraphics[width= 0.15\textwidth]{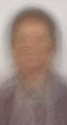} \\ \caption{Mean of original attacked class.} \label{fig:pp-means}
\end{figure}

While the outputs of the GAMIN attack against the CNN1 architecture remain blurry and noisy enough to disclose few features of the members of each class, the attack against the soft architecture gives away more precise contours. 
However, the attack against the Softmax skin classifier yields results which, unlike the ones obtained on MNIST classifier in Section~\ref{attack:results:results}, did not seem to converge to the attacked class mean (compare for instance Figure~\ref{fig:gamin-pp-skin-soft} to Figure~\ref{fig:pp-means}). 

In this case, it could be argue that even if the faces are undetermined, the clothes alone might be enough to constitute a privacy breach. 
The privacy leak depicted here is more similar to a property inference attack~\cite{ateniese2015hacking,ganju2018property}, which aim at inferring global properties of the training set of the target model. 
In contrast to previous implementations of such attacks~\cite{ateniese2015hacking,ganju2018property} that work in the white-box setting, GAMIN aims to infer properties of the target model's training set in the black-box agnostic setting. 

While Fredrikson and co-authors have already obtained recognizable faces against a model of similar architecture in the white-box setting \cite{Fredrikson15} and Tramèr and colleagues proposed a more efficient black-box attack based on successive extraction and inversion of the model \cite{Tramer16}, the GAMIN provides another black-box alternative that does not require the adversary to have information about the target's architecture.

\subsection{Cost of the attack}

The budget allowed for attacking each model was $1,280,000$ queries (or equivalently $20,000$ batches of $64$ queries). 
However, the protocol described previously in Section \ref{protocol} enabled to easily retrieve the required number of batches to achieve convergence in terms of the M-global score.
Depending on the attacked digits, the GAMIN achieves convergence towards ``final'' results after $2,000$ to $20,000$ batches of training, with an average of $2,500$ batches. 
In particular, each epoch requires $128$ queries to the target model, giving an average of $320,000$ queries per attack. 

Compared to other black-box attacks targeting even larger datasets ($\sim41,000$ online queries in~\cite{Tramer16}, and $20,600$ in~\cite{Fredrikson15} for attacking a model trained on the AT\&T Faces dataset of larger dimension~\cite{facescite}), the GAMIN is significantly more expensive in terms of queries. However, this increased cost could be attributed to the higher complexity of target models and is also balanced by the flexibility of GAMIN.

Despite the higher complexity in terms of queries, if we estimate that it takes $70$ ms in terms of latency for querying the target, as in~\cite{Fredrikson15,Papernot16} and not accounting for the speed increase offered by batch queries, \emph{a GAMIN attack could be completed in six hours}.
In addition assuming the adversary benefits from GPU acceleration, the networks training takes negligible time compared to querying the oracle.

\subsection{Attacks against models trained with differential privacy}

To evaluate how well our attack fare against protection mechanisms, we evaluate the effect of differential privacy on the success on an inversion attack against a MLP model.
For this experiment, we focus on the MLP1 architecture described in Section~4.1.3. 
To train models with this architecture with differential privacy, we use the noisy batched stochastic gradient descent algorithm~\cite{song2013stochastic}, with the implementation provided in \emph{tensorflow/privacy}~\footnote{https://github.com/tensorflow/privacy}. 
In particular, we set the privacy parameter $\delta$ to $1e^{-5}$ and train a model achieving a test accuracy of $0.89$ for a privacy parameter $\epsilon = 2.9$.

\begin{table}[h!]
\centering
$\begin{array}{ |*{3}{c|}|c| }
\hline 
\textnormal{Target model}  & \fidelity{}  & \accuracycomb{}  & M_{global}\\ \hline \hline
\textnormal{MLP1 without DP}          & 0.73 & 1.0       & 0.08  \\ \hline
\textnormal{MLP1 with DP}          & 0.74 & 0.96         & 0.16  \\ \hline
\end{array}$
\centering
\caption{Average performances across all digits after model inversion attacks against MLP models trained with (and without) differential privacy. $\fidelity{}$, $\accuracycomb{}$ and $M_{global}$ are respectively the fidelity of the surrogate, the categorical accuracy of the combined model, and the global convergence score.}
\label{tab:dpAttacks}
\end{table}

\begin{figure}[h!]
\centering
\includegraphics[width=0.5\textwidth]{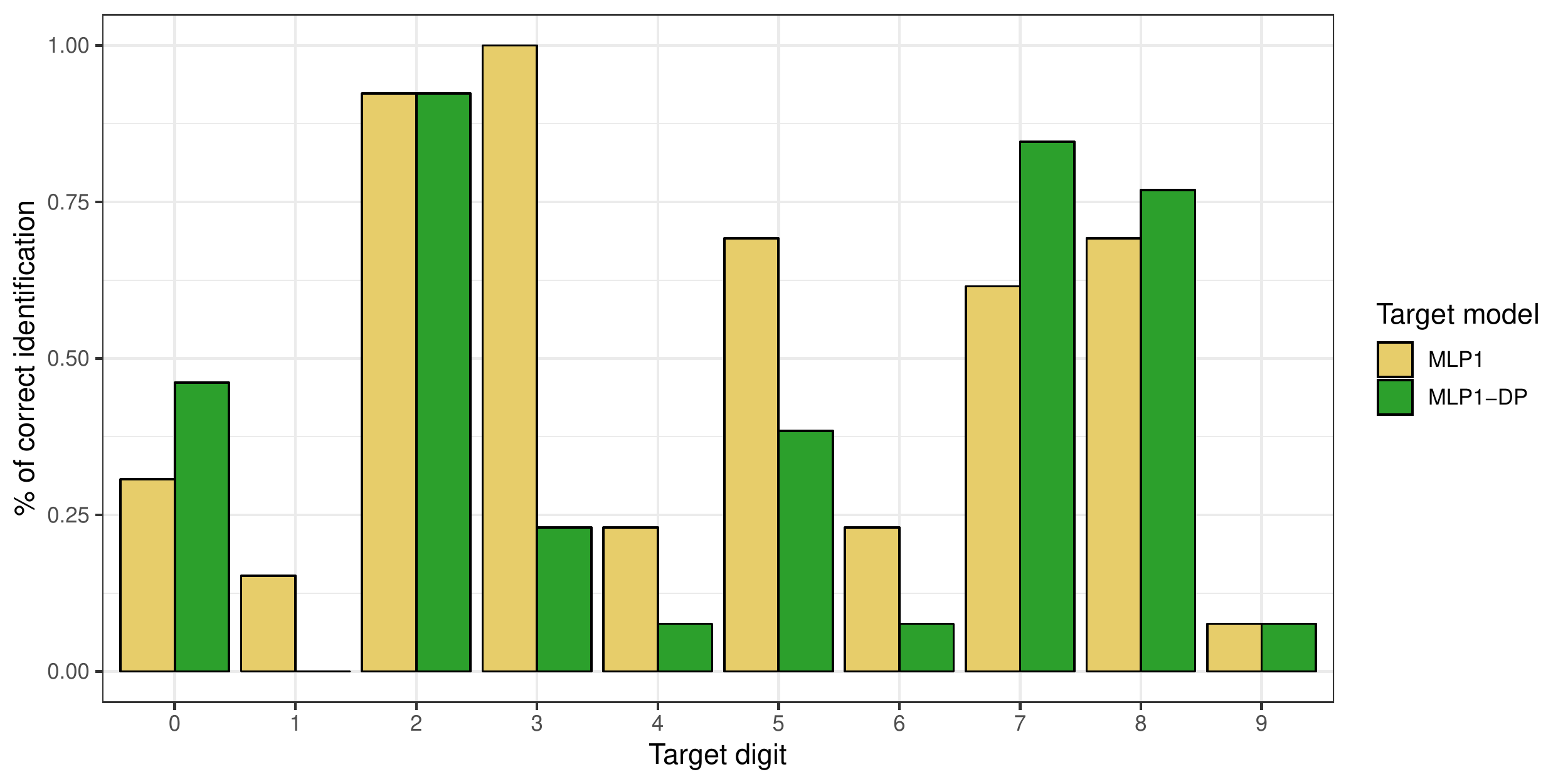}
\caption{Results of the survey for attacks against MLP1 model with (respectively without) differential privacy. For each target model and for each target digit, the proportion of correct identification is given.}
\label{fig:result:mnist:survey_dp}
\end{figure}

Table~\ref{tab:dpAttacks} and Figure~\ref{fig:result:mnist:survey_dp} summarize the results obtained. 
These results seem to indicate that the use of differential privacy does not prevent the convergence of the training of GAMIN and the success of the attack. 
This observation is consistent with findings from previous works that have analyzed the impact of differential privacy on inversion attacks~\cite{Hitaj17,abadi2017protection,jayaraman2019evaluating}. 
Intuitively this can be explained by the fact that differentially private learning techniques are essentially designed to protect against membership attack. 
In contrast protecting against model inversion attacks is similar in spirit as protecting the membership of all items that belong to a particular class, which require to apply some notion of group privacy. 
However, by doing so, the utility of the model will be negatively impacted.

\section{Related work}
\label{sec:relatedWork}

Prior works related to privacy attacks against discriminative models in the supervised learning setting include adversarial attacks, membership inference attacks~\cite{Shokri17, melis2018exploiting}, property inference attacks~\cite{ateniese2015hacking,ganju2018property}, model inversion attacks~\cite{Fredrikson15, Hitaj17} and model extraction attacks~\cite{Tramer16, wang2018stealing, milli2018model}. 
In addition to regularization techniques, differential privacy is the workhorse used to address most of the inference attacks against machine learning models.

\subsection{Adversarial attacks}

The main idea of the adversarial setting is that an attack consists in crafting an input, which is then sent to the target model. 
The objective is to affect the target model in a controlled and desired manner. 
For example, \emph{evasion attacks} aims to avoid a certain classification,  in the sense that the objective is to produce anything but the specified class for a certain input \cite{Biggio,Laskov2009}.
Such attacks were successfully performed against shallow models such as SVMs or RBF Kernels, using a gradient-descent-based algorithm, by crafting adversarial examples that crossed decision boundaries by a large margin (which translates as a high confidence in the faked class) \cite{Biggio}. 
A more modern approach based on GANs has also proved to be able to fool deep neural models \cite{Papernot16}, even on temporal or recurrent architectures \cite{Papernot16:recurrent}.

Both of the approaches described above rely on learning how to respectively maximize a utility or minimize a loss function reflecting the confusion of the target model. 
The learning process uses the gradient of this function to change parameters but such gradients are not accessible in a white-box setting.
However, this can be countered by the concept of \emph{shadow training} \cite{Shokri17}, which consists in training surrogate models, learning to attack these surrogate - or ``shadow'' - models and then transferring the attack to the target model \cite{Tramer17:transferable,Tramer17:atkdef}.
The use of a surrogate model in the GAMIN is inspired from the \emph{shadow training} method.

\subsection{Inference attacks against machine learning models} 
\label{sec:inferences}

Membership attacks against machine learning models have been introduced by Shokri, Stronati, Song and Shmatikov~\cite{Shokri17}. 
Given a data record $d$ and a trained model $M$ trained over a training dataset $D^{M}_{train}$, a membership inference attack consist in trying to evaluate if $d \in D^{M}_{train}$.
For instance, the authors demonstrated in 2017 the possibility for an adversary to assess the presence of a given individual in hospital datasets in a true black-box setting, highlighting the potential privacy damage this type of attack can cause.
This type of attack exploits the fact that machine learning models may be subject to overfitting (\emph{i.e,} being significantly more accurate at predicting outputs for the training data than predicting outputs for the test data). 
The attack involves training multiple shadow models, each using the same machine learning technique as that of the target model, and using a dataset similar to that of the target model.
However, this is done by explicitly labeling predictions vectors on its training set and its test set differently. 
Finally, a classifier is trained to distinguish training data from test data. Membership attacks have also been studied by Melis, Song, de Cristofaro and Shmatikov~\cite{melis2018exploiting} in the context of collaborative learning, in which the authors showed that the interactive nature of the collaboration can be exploited by a participant to conduct a membership attack on other participants' training sets.
In addition, Hayes, Melis, Danezis and de Cristofaro have demonstrated in the context of generative models \cite{hayes2018logan} that generative adversarial networks \cite{gan-Goodfellow2014} can be used to infer the presence of a particular individual in the training set.

Property inference attacks against machine learning models have been introduced by Ateniese and co-authors~\cite{ateniese2015hacking}. 
This type of attack involves training a meta-classifier to detect if the target model has a given property $P$. 
To conduct such an attack, the adversary trains a set of shadow models using a dataset and machine learning technique similar to that of the target model, but in addition, explicitly labeled as having the property $P$ or not. 
Finally, the meta-classifier is trained to detect the presence of the property $P$. 
The authors have used this attack to learn that the training set of a speech recognition system have been produced by people speaking a particular dialect.
Remark that the shadow training technique developed here is the same that has inspired the membership attacks \cite{Shokri17}. 
More recently, property inference attacks have been studied~\cite{melis2018exploiting} in the context of collaborative learning, in which the authors demonstrated that the collaborative gradient update that occurs in this type of learning can be exploited by the adversary to infer properties that are true for a subset of the training set of other participants.
Ganju, Wang, Yang, Gunther and Borisov~\cite{ganju2018property} have further demonstrated the effectiveness of property inference attacks on more complex machine learning models such as fully connected neural networks. 

Model inversion attacks aim at inferring, given a target model $f$ and an output class $y$, sensitive hidden features of inputs corresponding to the class $y$. 
As a result, the adversary will learn the average of the features of inputs that belong to the class $y$. 
Model inversion attacks have been used originally~\cite{Fredrikson15} to infer if participants to a survey admitted of having cheated on their significant other by inverting decision trees, and to reconstruct people's faces by inverting a facial recognition system trained using three different classifiers, namely a softmax classifier, a multilayer perceptron network and a denoising autoencoder. 
Recently, Hitaj, Ateniese and Perez-Cruz~\cite{Hitaj17} have demonstrated that model inversion can be more powerful in collaborative machine learning context in which the adversary can train a generative adversarial network during the update phase to create prototypical examples of its target's training set. 
Hidano, Murakami, Katsumata, Kiyomoto and Hanaoka~\cite{hidano2018model} have introduced a model inversion attack that does not leverage on the knowledge of non-sensitive attributes of the input, under the assumption that the target model is operating in the online setting. 
In particular, a poisoning attack~\cite{biggio2012poisoning} is conducted on the target model to turn the regression coefficients of the non-sensitive attributes to zero. 
The resulting target model then becomes easy to invert, using the same approach proposed in~\cite{fredrikson2014privacy,Fredrikson15}, without requiring additional knowledge of the non-sensitive attributes.

Finally, model extraction attacks aim at inferring, given a target model $f$ and its predictions (or explanations) for a chosen set of inputs, the parameters and/or hyper-parameters of $f$. 
Tramer and co-authors~\cite{Tramer16} have demonstrated the effectiveness of such attacks by reconstructing several machine learning models after querying online machine learning as a service platforms. 
Wang and Cong~\cite{wang2018stealing} have also proposed attacks to steal hyper-parameters of several machine learning models. 
More recently, Milli, Schmidt, Dragan and Hardt~\cite{milli2018model} have shown how attacks exploiting gradient-based explanations can allow to extract models with significant less query budget compared to attacks based on predictions. 
Batina, Bashin, Jap and Picek~\cite{batina2018csi} have proposed an attack against neural networks that leverages on a side-channel attack. In particular, it monitors the power use of the microprocessor on which the target model is evaluated to extract its weights. 
Jagielski, Carlini, Berthelot, Kurakin and Papernot~\cite{jagielski2019high} have leveraged on MixMatch~\cite{berthelot2019mixmatch}, a semi-supervised learning technique, to produce a query-efficient learning-based extraction attack, which trains a surrogate on inputs labelled by the target model. 
The authors also introduced a functionally-equivalent extraction attack, which produces a surrogate that belongs to the equivalence class of the target model, that demonstrated better fidelity. 
Finally, they proposed a hybrid approach that combines functionally-equivalent extraction attacks and learning-based extraction attacks to improve the fidelity of the surrogate model.

Remark that model extraction is not per se a privacy attack. 
However, it can be used to build white-box surrogates of the target model and thus ultimately help in improving the efficiency of other inference attacks.
The main difference between GAMIN and the existing attacks is that it operates in a black-box agnostic setting, which allows the attack to be easily deployed against any machine learning model. 
In addition, the computational cost of the attack (about 6 hours on a MLP) is acceptable. 
To compare, the estimated cost of a black-box attack --- based on numeric gradient approximation~\cite{Fredrikson15} --- against a MLP is about $50$ to $80$ days.

\subsection{Countermeasures against inference attacks}

Adding noise to individual records while maintaining global distribution and correlations is at the core of the concept of differential privacy, a privacy model developed by Dwork and co-authors \cite{Dwork13} to balance the risk of privacy loss and the degradation of accuracy results of analysis run on a dataset of private records.
Recently, the application of differential private methods to machine learning has drawn a lot of attention from privacy researchers and stemmed multiple approaches.
One of these approaches relied on the training of deep networks over data distributed among multiple users \cite{Shokri15}. 
Afterwards, new attacks were designed to address this specific setting \cite{Hitaj17, Melis18}. 
Differential privacy was also integrated in the design of a stochastic gradient descent algorithm \cite{Abadi16}. 
However, the inherent trade-off between the privacy level and the performance of models trained with this method has already been pointed out in previous works~\cite{attacks against DP,jayaraman2019evaluating}. 

To measure the resistance of a model against membership inference attacks, Long, Bindschaedler and Gunter introduced the concept of differential training privacy~\cite{DTP} and showed that countermeasures such as \emph{distillation} may not be sufficient to counter this type of attack.
In their study of model inversion attacks~\cite{Fredrikson15}, Fredrikson, Jha and Ristenpart suggest to round confidence scores to reduce the vulnerability of models to equation-solving attacks in the black-box setting. However, this can decrease the precision on the gradient retrieved from target model as intended. 
Against a black-box GAMIN attack, this would reduce the capacity of the surrogate to correctly approximate the target's decision boundaries. 
While this may result in a lesser quality of the reconstructions or even a failure of GAMIN attacks, the efficiency of this defense could be also differ greatly as boundaries retrieved by the surrogate could appear less strict, which could ultimately facilitate the training of the generator.

\section{Conclusion and future works}
\label{sec:conclusion}

In this paper, we have demonstrated how adversarial approaches can be adapted to achieve model inversion attacks against machine learning models in an agnostic black-box setting. 
We have shown the potential threat of privacy leakage posed by these attacks by testing our architecture against models trained on a dataset of handwritten digits and another of face portraits for which we were able to reconstruct recognizable digits or features.
In addition our attack works in multiple scenarios, including the situation in which the adversary has absolutely no prior knowledge of the target models and without specific parameter optimization.

We have also reviewed a variety of measures that could be proposed to measure the success of model inversion attacks and shows that usual and intuitive metrics may not be sufficient to predict the attack success. 
Furthermore, we show that models of increasing complexity, like convolutional neural networks, are more resistant to model inversion thanks to the abstraction and dilution of their decision boundaries.

We are planning to extend our approach in several directions. 
In particular, future work will include the testing of GAMIN against different types of targets, such as regressors or recurrent architectures and the study of countermeasures against this type of attack. 
We believe that the GAMIN can still be improved to further improve accuracy or to decrease the cost of this attack in terms of queries.
The GAMIN could benefit from being tested against targets trained on different types of data, either as inputs (\emph{i.e.}, sequential data for instance) or as outputs (\emph{i.e.}, regressors).
Further development of this architecture could allow for instance a single GAMIN to invert multiple outputs. 
In this case, the cost of the attack would not decrease directly, but a single attack could invert multiple classes, which would dramatically reduce the cost of inverting all classes of the model.

\clearpage

\appendix
\section{Gallery of models}
\label{anx:galleryOfModels}
This appendix lists the models used in this study, both as parts of the GAMIN architecture and as targets models.

\subsection{Image classification}

\subsubsection{GAMIN surrogate model}\label{anx:archi:gamin:sur}

\begin{table}[h!]
\centering
\begin{tabular}{c|c|c|c|c|c|c|c|c}
IN  & OUT & CM & CM & RL  & D   & RL  & D   & S  \\ \hline
784 & 10  & 32 & 64 & 128 & 0.5 & 32  & 0.5 & 10 \\  
\end{tabular}
\caption{GAMIN surrogate for MNIST image classification model attacks. 
IN denotes the dimension of inputs, OUT the dimension of outputs, CM is a 2D convolution of kernel size (3x3) with ReLU activation followed by a max pooling over (2x2), RL is a Rectified Linear Unit fully-connected layer, D is the proportion of dropout between these layers and S a softmax (output) layer.}\label{tab:surrogate}
\end{table}

Table \ref{tab:surrogate} describes the model used to perform the GAMIN attack against models trained over the MNIST dataset. 
Note that since the convolutions used in this model have a different filter size than those used for the targets (see Section \ref{anx:targets}), the surrogate has very little in common with the target beyond the dimensions of its inputs and outputs.

\subsubsection{Target models} \label{anx:targets}

\begin{table}[h!]
\centering
\begin{tabular}{c|c|c|c|c|c|c|c}
Name & IN  & OUT & CM & CM & RL  & RL  & S  \\ \hline
Soft & 784 & 10  &    &    &     &     & 10 \\ \hline
MLP1 & 784 & 10  &    &    & 100 &     & 10 \\ \hline
MLP2 & 784 & 10  &    &    & 200 & 200 & 10 \\ \hline
CNN1 & 784 & 10  & 32 &    & 200 &     & 10 \\ \hline
CNN2 & 784 & 10  & 32 & 64 & 200 &     & 10 \\ 
\end{tabular}
\caption{Target models for MNIST image classification. IN denotes the dimension of inputs, OUT the dimension of outputs, CM is a 2D convolution of kernel size (2x2) with ReLU activation followed by a max pooling over (2x2), RL us a Rectified Linear Unit fully-connected layer and S is a softmax (output) layer.}\label{tab:targets}
\end{table}

Table \ref{tab:targets} describes all models used to classify MNIST digits.

\subsection{Image generation}

\subsubsection{GAMIN generator mode}\label{anx:archi:gamin:gen}

\begin{table}[h!]
\centering
\begin{tabular}{c|c|c|c|c|c|c|c|c}
IN  & OUT & L   & CM  & D   & CM  & D   & CM & TAN  \\ \hline
10 & 784  & 784 & 128 & 0.5 & 128 & 0.5 & 64 & 784 \\  
\end{tabular}
\caption{GAMIN Generator for MNIST image classification model attacks. IN denotes the dimension of inputs, OUT the dimension of outputs, CM is a 2D convolution of kernel size (3x3) with ReLU activation followed by a max pooling over (2x2), D is the proportion of dropout between these layers and TAN is a fully-connected layer with $tanh$ activation (output layer).}\label{tab:generator}
\end{table}

Table \ref{tab:generator} displays the details of the GAMIN Generator architecture for reconstructing handwritten digits.
\section{Target performance breakdown}
\label{anx:target-performance}

All the accuracies of target models are listed in Tables \ref{tab:targets-perf} and \ref{tab:targets-perf-pp} for models trained respectively, on the MNIST and on the pilot parliament datasets. We denote by $A_{train}$ \& $A_{test}$ the accuracy respectively over the training and testing datasets.

\begin{table}[h!]
\centering
$\begin{array}{ |*{3}{c|} }
\hline 
\textnormal{MNIST target digit} & A_{train} & A_{test} \\ \hline \hline
\textnormal{Soft} & 0.85 & 0.83 \\ \hline
\textnormal{MLP1} & 0.93 & 0.91 \\ \hline
\textnormal{MLP2} & 0.97 & 0.94 \\ \hline
\textnormal{CNN1} & 0.99 & 0.99 \\ \hline
\textnormal{CNN2} & 0.98 & 0.99 \\ \hline
\textnormal{Random baseline} & 0.1 & 0.1 \\ \hline
\end{array}$
\caption{Target accuracy on MNIST Dataset}\label{tab:targets-perf}
\end{table}

\begin{table}[h!]
\centering
$\begin{array}{ |*{3}{c|} }
\hline 
\textnormal{Pilot Parliament: skin color} & A_{train} & A_{test} \\ \hline \hline
\textnormal{Soft} & 0.43 & 0.33 \\ \hline
\textnormal{CNN1} & 1.0 & 0.99 \\ \hline
\textnormal{CNN2} & 1.0 & 1.0 \\ \hline
\textnormal{Random baseline} & 0.17 & 0.17 \\ \hline
\end{array}$
\caption{Target accuracy on pilot parliament dataset.}\label{tab:targets-perf-pp}
\end{table}

\end{document}